\documentclass[10pt,twocolumn,letterpaper]{article}

\usepackage{iccv,times,graphicx,tabulary,multirow,xspace,xcolor}
\usepackage{amsthm,amssymb,fixmath,mathtools,nicefrac}
\usepackage[pagebackref=true,breaklinks=true,letterpaper=true,colorlinks,citecolor=citecolor,bookmarks=false]{hyperref}
\definecolor{citecolor}{RGB}{34,139,34}

\usepackage{amsmath}
\usepackage{pifont}

\usepackage{caption}
\usepackage{subcaption}

\begin{document}
\title{Equalization Loss for Large Vocabulary Instance Segmentation}
\track{LVIS}

\author{Jingru Tan\\
Tongji University\\
{\tt\small tjr120@tongji.edu.cn}
\and
Changbao Wang\\
SenseTime Research\\
{\tt\small wangchangbao@sensetime.com}
\and
Quanquan Li\\
SenseTime Research \\
{\tt\small liquanquan@sensetime.com}
\and
Junjie Yan\\
SenseTime Research \\
{\tt\small yanjunjie@sensetime.com}
}

\maketitle

\begin{abstract}
Recent object detection and instance segmentation tasks mainly focus on datasets with a relatively small set of categories, e.g. Pascal VOC with 20 classes and COCO with 80 classes.
The new large vocabulary dataset LVIS brings new challenges to conventional methods.
In this work, we propose an equalization loss to solve the long tail of rare categories problem.
Combined with exploiting the data from detection datasets to alleviate the effect of missing-annotation problems during the training, our method achieves 5.1\% overall AP gain and 11.4\% AP gain of rare categories on LVIS benchmark without any bells and whistles compared to Mask R-CNN baseline.
Finally we achieve 28.9 mask AP on the test-set of the LVIS and rank 1st place in LVIS Challenge 2019.
\end{abstract}


\section{Introduction}
Different from preceding instance segmentation datasets such as COCO~\cite{lin2014microsoft}, the large vocabulary instance segmentation dataset LVIS~\cite{gupta2019lvisarvix} poses new challenges.
First, unbalanced data distribution of categories leads to serious performance degradation of rare categories.
Second, LVIS is not exhaustively annotated with all categories therefore unlabeled object instances will be treated as background and will generate incorrect supervision signals.
Recent state-of-the-art methods show poor performance on LVIS~\cite{gupta2019lvisarvix}, especially for the rare categories.
In this work, we focus on these two problems. 
For the long-tail problem, we propose a new loss function to improve the performance of rare categories, which will be described in Section~\ref{EQL}.
For the missing-annotation problem, we provide simple but effective strategies to utilize object detection data and annotations, which will be described in Section~\ref{data}.


\begin{figure}[ht]
\begin{subfigure}{.23\textwidth}
  \centering
  \includegraphics[width=1.1\linewidth]{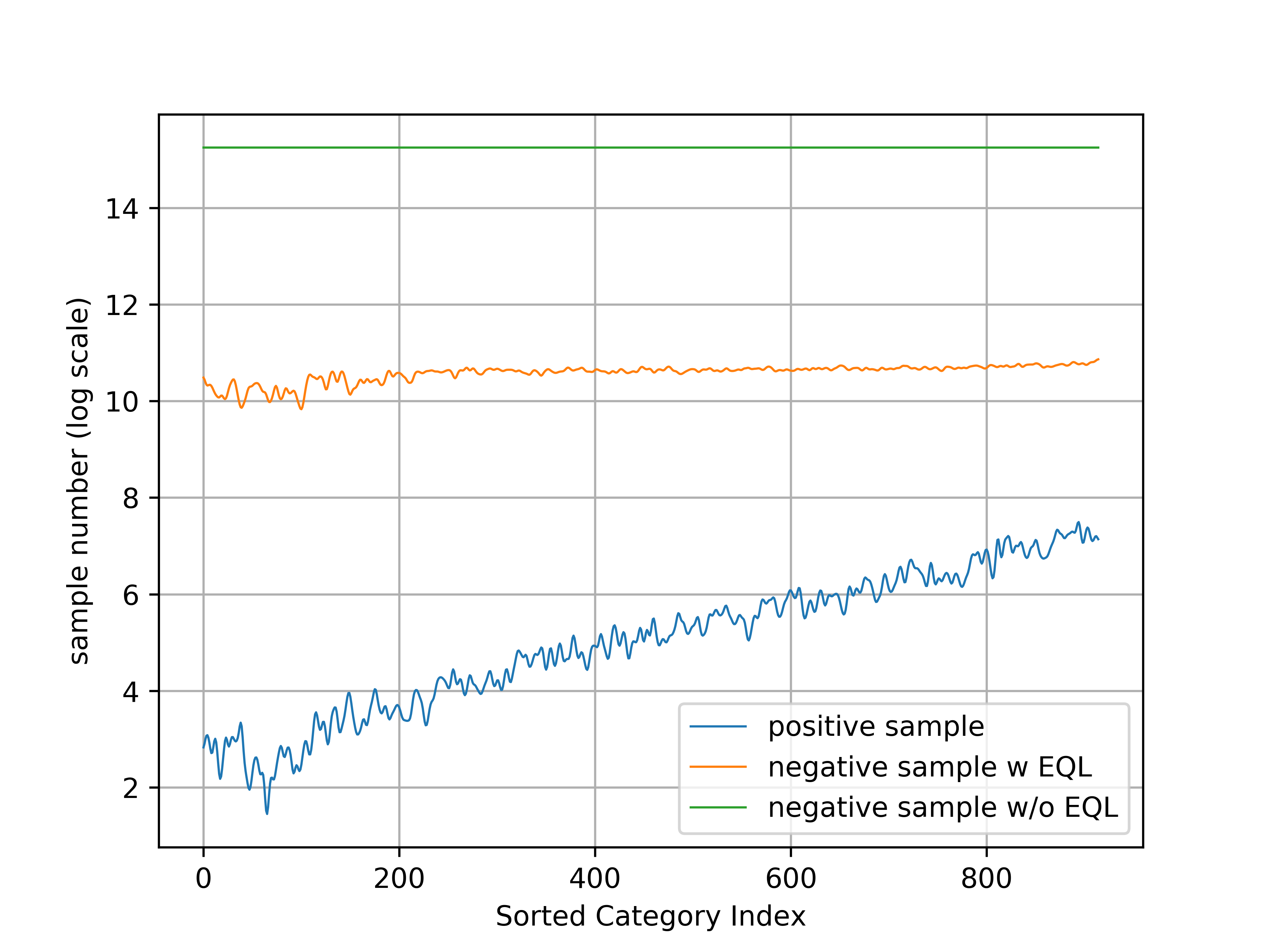}  
  \caption{}
  \label{fig:sub-first}
\end{subfigure}
\begin{subfigure}{.23\textwidth}
  \centering
  \includegraphics[width=1.1\linewidth]{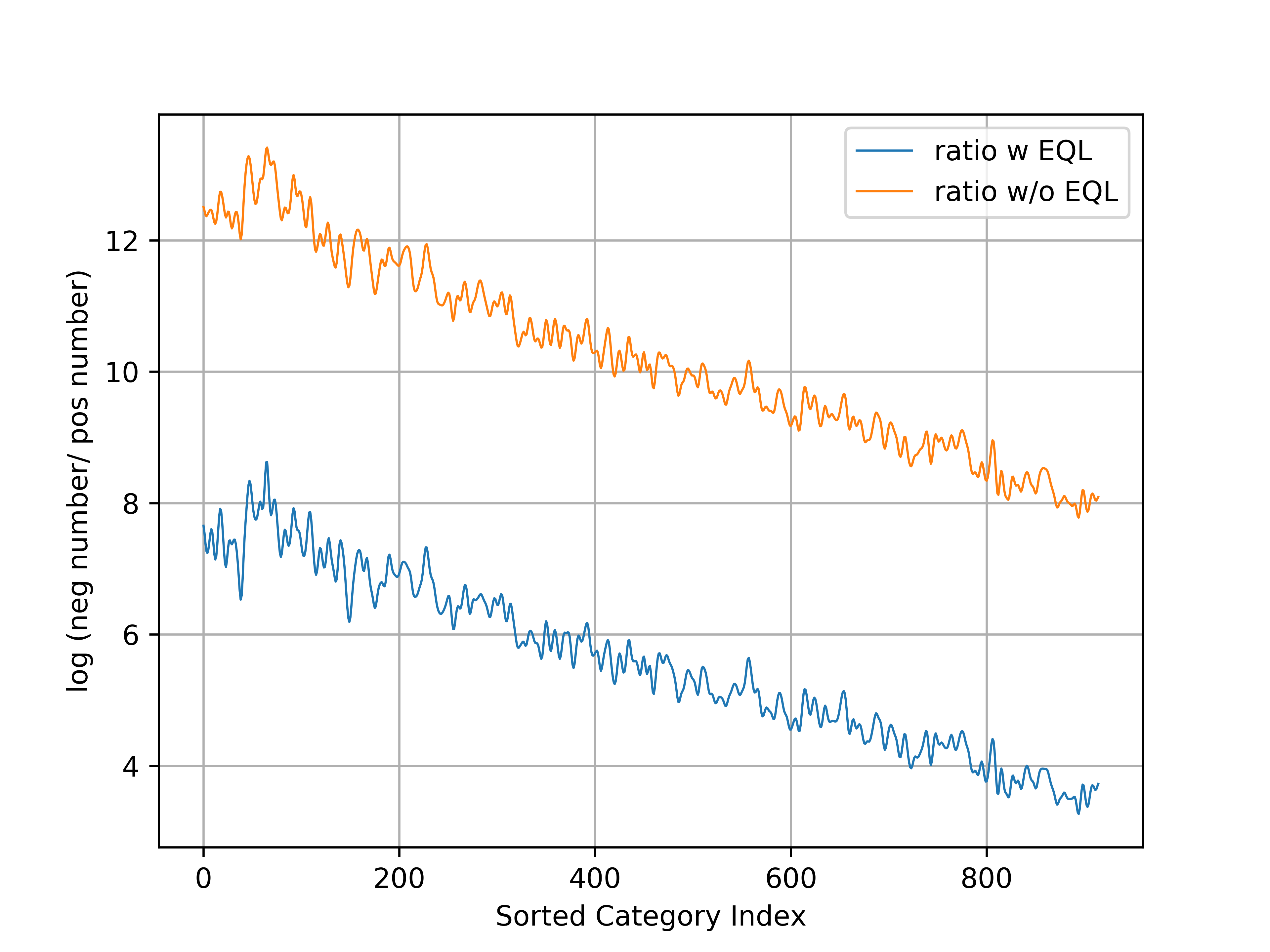}  
  \caption{}
  \label{fig:sub-second}
\end{subfigure}
\caption{The effect of equalization loss on valid positive and negative samples ("valid" means samples which have contributions to the loss). (a) shows the number of valid positive and negative samples for each category; (b) displays the ratio of the number of negative samples to the number of valid positive samples. Categories are sorted by their frequency.}
\label{fig:effect_of_eql}
\end{figure}

\section{Equalization Loss} \label{EQL}
Our work is based on the state-of-the-art instance segmentation framework Mask R-CNN \cite{he2017mask} with two modifications.
First, to alleviate the competition between categories, we replace softmax cross-entropy loss with sigmoid cross-entropy loss for the box classification.
Second, to reduce the computation and memory cost, we use class-agnostic mask prediction for the mask head instead of class-specific mask prediction in origin paper~\cite{gupta2019lvisarvix}.

During the box prediction stage of Mask R-CNN, for proposal $R$ assigned with category $c$, the sigmoid cross-entropy loss of classification branch can be computed as:

\begin{equation}
    L_{cls} = -\sum_{j=1}^{C} log(p_{j}^{*}),
    \label{equation1}
\end{equation}
which
\begin{equation}
    p_{j}^{*} =
    \begin{cases}
        p_{j} & \text{if } j = c \\
        1 - p_{j}              & \text{otherwise,}
    \end{cases} 
    \label{equation2}
\end{equation}


\noindent where $C$ is the total number of categories, $p_{j}$ is the predicted confidence for category $j$. 
This loss function requires that for a given proposal, it should try to predict only one category. 
However, in LVIS, one object can be annotated with multiple categories, there is no strict boundary between some categories. Meanwhile, since the annotations of rare and common categories are much less than that of frequent categories, predictions for rare and common categories are suppressed for almost all the time using Equation \ref{equation1} and  \ref{equation2}. In other word, a positive sample of one category can be seen as a negative sample for other categories at the same time.
Those negative signals have a marked impact on categories with scarce annotations, i.e.\ rare and common.
We claim that less punishment to the rare the common objects helps alleviate the two problems mentioned above, so we introduce a novel equalization loss. It introduce an additional weight $w \in \mathbb{R}^{C}$ to the origin sigmoid loss function. Given a proposal $R$, we compute $w$ as follows: if the proposal $R$ is negative, $w$ is set to 1 for all index $j$; For a positive $R$, $w$ is set to 0 if $j$ $<$ $\lambda$ and its category $c$ is not in the union of positive category set and negative category set. The equalization loss is formulated as:

\begin{equation}
    L_{EQL} = -\sum_{j=1}^{C} w_{j} log(p_{j}^{*}),
\end{equation}
which 
\begin{equation}
    w_{j} =
    \begin{cases}
        0 & \text{if } c > 0 \text{ and } f_{j} < \lambda \text{ and } j \notin S_{P} \cup S_{N} \\
        1 & \text{otherwise}
    \end{cases} 
\end{equation}

\noindent where $c$ is the category for proposal $R$, $f_{j}$ is the frequency of category $j$, $S_{P}$ and $S_{N}$ are the positive and negative category sets of the ground truth annotations of the image. 
Since the categories are classified to frequent, common and rare in LVIS, in our experiments, we empirically set $\lambda$ to ignore all rare and common categories.
A more detailed parameter search of $\lambda$ may bring further improvements. 

The effect of the proposed loss is shown in Figure \ref{fig:effect_of_eql}. As we can see, it alleviates the imbalance problem between positive and negative samples.

\begin{figure}[ht]
\begin{subfigure}{.23\textwidth}
  \centering
  \includegraphics[width=\linewidth]{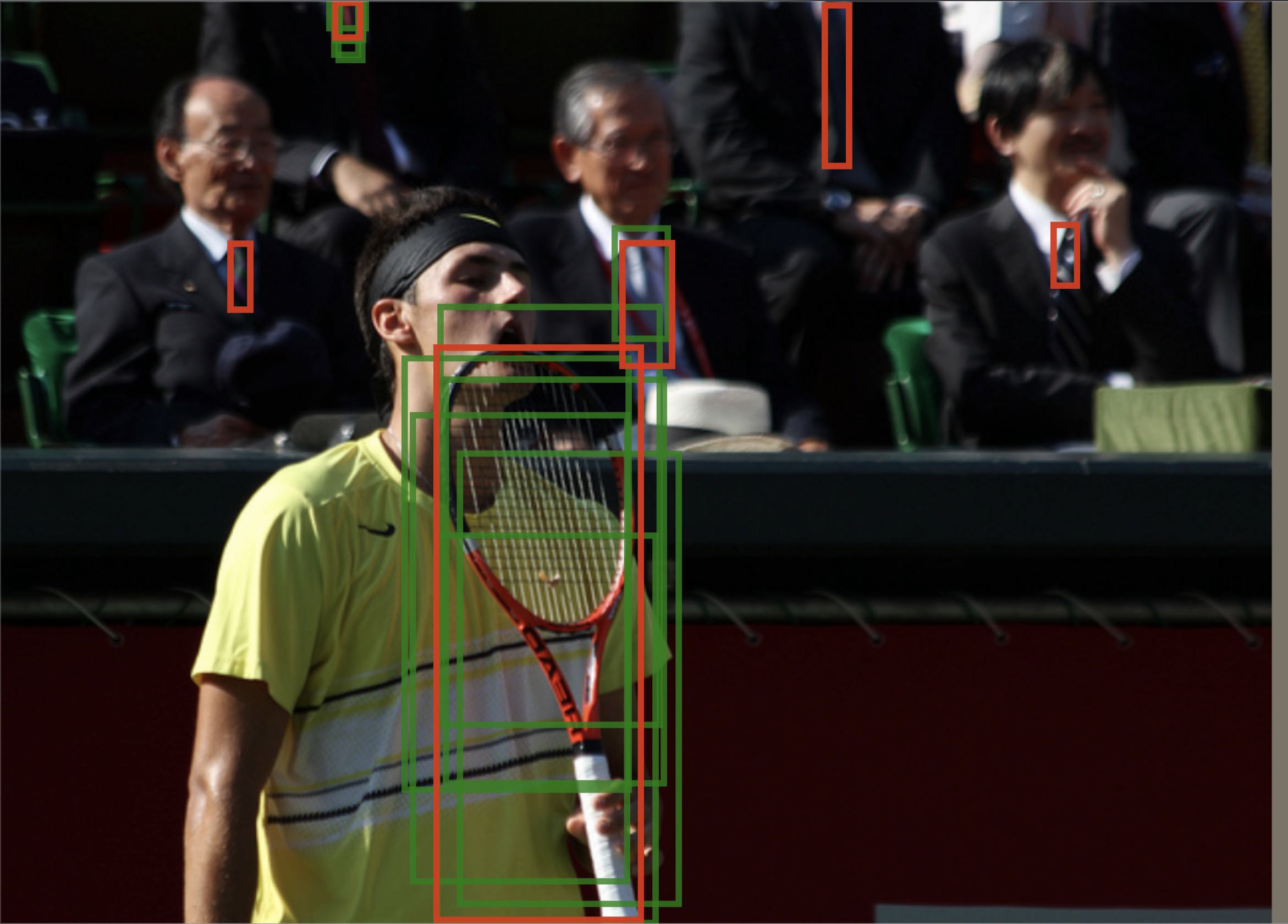}  
  \label{fig:sub-first}
\end{subfigure}
\begin{subfigure}{.23\textwidth}
  \centering
  \includegraphics[width=\linewidth]{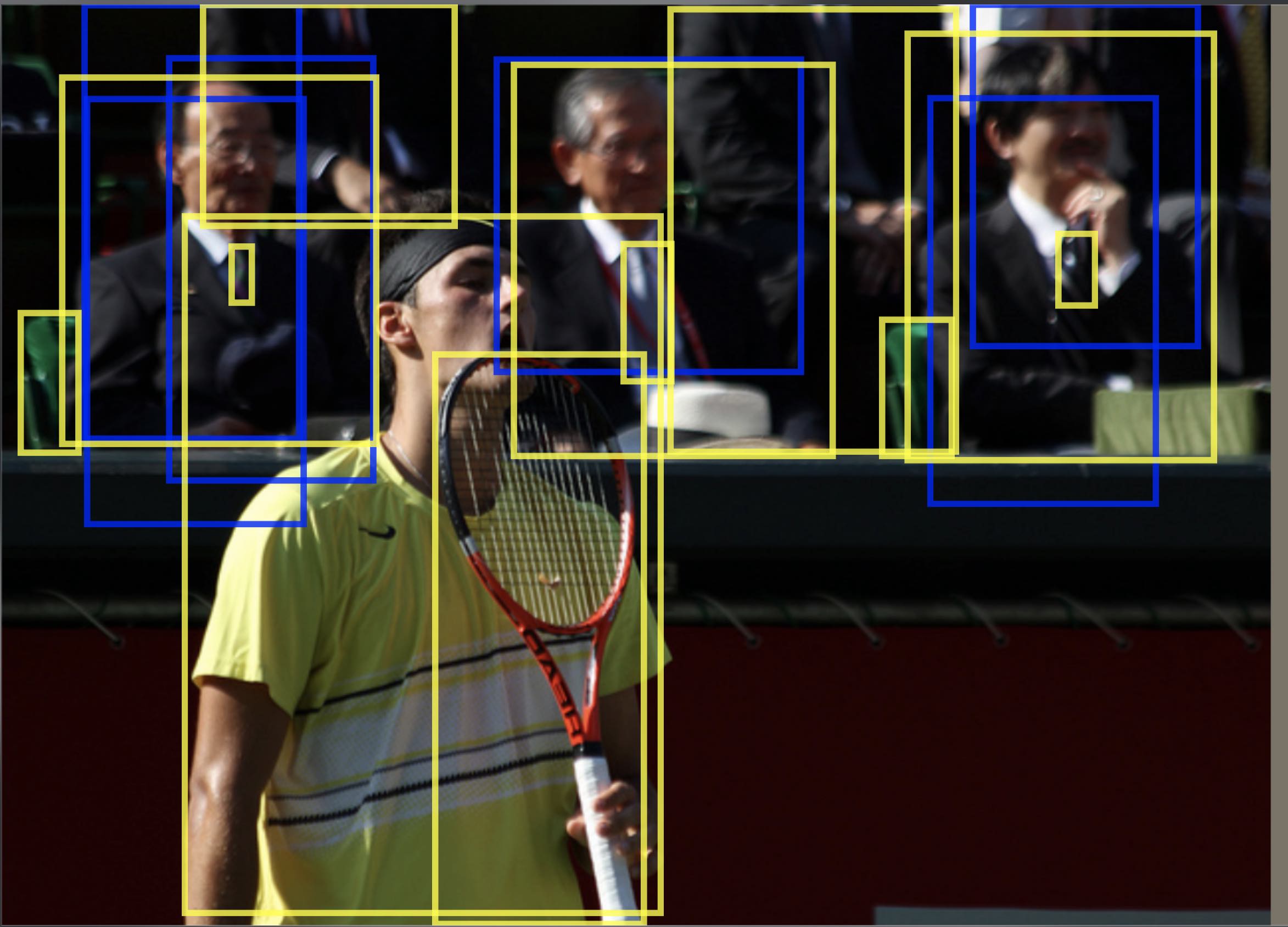}  
  \label{fig:sub-second}
\end{subfigure}
\caption{Examples of miss-annotation. Red and green boxes are LVIS annotated boxes and positive proposals, yellow and blue boxes are COCO annotated boxes and negative proposals. Those blue negative proposals will give incorrect updating signals to the model. }
\label{fig:coco_ignore}
\end{figure}

\section{Exploiting Data of Object Detection} \label{data}
Since LVIS is not exhaustively annotated with all categories and data of rare categories are quite scarce, we utilize additional public datasets and provide several strategies.
\subsection{COCO Ignore}
If a proposal is assigned to negative label because of missing-annotation, (miss-annotations problem comes from unknown categories and "not exhaustive" cases, not annotations errors), the model will get an incorrect updating signal, which will influence the training and degrades the performance. 
Since LVIS and COCO share the same set of images, we utilize the bounding box annotation from COCO dataset.
During training, we calculate the overlaps between negative proposals and COCO ground truth bounding boxes. 
Figure \ref{fig:coco_ignore} shows some examples of the missing-annotation situation.
For those IoU larger than 0.5, we decrease the weight to $\beta$. 
Since this strategy changes the total loss scale, we double the loss weight of the box head and set $\beta$ to 0.5 rather than 0.

\subsection{COCO Pre-training}
Transfer learning is helpful for a relatively small dataset, and LVIS shares the same image sets with COCO, so it's intuitive to train a detector in LVIS using COCO pre-trained model instead of ImageNet pre-trained model.
We pre-train our model on COCO with instance segmentation annotations and then fine-tune on LVIS.

\subsection{OpenImage}
OpenImage~\cite{kuznetsova2018open} is a large datasets with 600 object categories. 
LVIS shares 110 categories with OpenImage. 
We add corresponding (about 20k images) images to LVIS train set. 
Only bounding-box level annotations are used, so the losses of mask branch are ignored for those images of OpenImage dataset.

\begin{table}
    \centering
    \begin{tabular}{l|c c c c}
        Model & $\mathrm{AP}$ & $\mathrm{AP_{r}}$ & $\mathrm{AP_{c}}$ & $\mathrm{AP_{f}}$  \\
        \hline
        ResNet50-Softmax \cite{gupta2019lvisarvix} & \textbf{21.0} & 3.2 & 21.3 & 27.7 \\
        \hline
        ResNet50-Softmax\textsuperscript{*} & 20.8 & 5.6 & \textbf{21.5} & 25.6 \\
        ResNet50-Sigmoid & 20.1 & \textbf{6.5} & 19.9 & 25.4 \\
         \hline
    \end{tabular}
    \caption{Baseline model results on LVIS val v0.5. ResNet50-Softmax\textsuperscript{*} is our re-implementation. The metrics are mask AP and subscripts 'r', 'c', and 'f' stands for rare, common, and frequent category.}
    \label{tab:baseline}
\end{table}

\begin{table}
    \centering
    \begin{tabular}{l|c c c c}
        Model & $\mathrm{AP}$ & $\mathrm{AP_{r}}$ & $\mathrm{AP_{c}}$ & $\mathrm{AP_{f}}$  \\
        \hline
        Baseline Sigmoid Loss & 20.1 & 6.5 & 19.9 & 25.4 \\
        \hline 
        Equalization Loss & 22.8 & 10.1 & 25.0 & 25.1 \\
         \hline
    \end{tabular}
    \caption{Comparison of Equalization Loss and naive sigmoid loss. All model are trained using ResNet-50 Mask R-CNN.}
    \label{tab:eql}
\end{table}


\begin{table}
    \centering
    \begin{tabular}{c c c c|c c c c}
         EQL & RS & PR & IG & $\mathrm{AP}$ & $\mathrm{AP_{r}}$ & $\mathrm{AP_{c}}$ & $\mathrm{AP_{f}}$  \\
        \hline
         & & & & 20.1 & 6.5 & 19.9 & 25.4 \\
        \hline
         & \ding{52} & & & 21.3 & 12.2 & 21.5 & 24.7 \\ 
        \ding{52} & & & & 22.8 & 10.1 & 25.0 & 25.1 \\
        \ding{52} & \ding{52} & & & 23.3 & 15.3 & 24.8 & 24.5 \\
        \ding{52} & & \ding{52} &  & 23.9 & 11.7 & 26.0 & 26.1 \\
        \ding{52} & & & \ding{52} & 23.5 & 14.8 & 25.0 & 25.2 \\
        \ding{52} & \ding{52} & \ding{52} &  & \textbf{25.2} & \textbf{17.9} & \textbf{26.8} & \textbf{26.2} \\
         \hline
    \end{tabular}
    \caption{Experiment results of \textbf{EQL}(Equalization Loss), \textbf{RS}(Resampling), \textbf{PR}(COCO Pretrain), \textbf{IG}(COCO Ignore)}
    \label{tab:ablation}
\end{table}

\section{Experiments}

We perform experiments on LVIS dataset, which contains 1230 categories in release v0.5.  
The evaluation metric is AP across IoU threshold from 0.5 to 0.95. 
We train our model on 60k \texttt{train} images and test it on 5k \texttt{val} set. 
We also reported our results on 20k \texttt{test} images.





\subsection{Implementation Details}

We implement standard Mask R-CNN equipped with FPN \cite{lin2017feature} as our model. 
Training images are resized such as its shorter edge is 800 pixels. 
No other augmentation is used except horizontal flipping. RPN samples 256 anchors with 1:1 ratio of positive to negative. 
RoIAlign \cite{he2017mask} is adopted to extracted features of proposals. 
R-CNN head samples 512 proposals per image,  with 1:3 ratio of  positive to negative. 
Though class-specific mask prediction achieves better performance, we use a class-agnostic regime due to memory and computation cost for mask branch.
In testing, the score threshold is reduced from 0.05 to 0.0, and top 300 bounding boxes are kept as detection results. 
Other settings are kept the same as origin implementation if not mentioned. 

\subsection{Ablation Study}
In this section, we perform ablation studies among Equalization Loss, COCO Ignore, COCO pre-training and class-aware resampling. 
We implement Mask R-CNN with ResNet-50 and replace the conventional softmax cross-entropy loss with sigmoid cross-entropy loss in box head as our baseline.
Comparisons of sigmoid loss and softmax loss used in origin paper~\cite{gupta2019lvisarvix} are shown in Table~\ref{tab:baseline}.

\noindent \textbf{Equalization Loss} The experiments results are shown in Table \ref{tab:eql}. 
Comparing with the sigmoid cross-entropy loss, our method can lead to a significant improvement from 20.1\% to 22.8\%, especially on rare and common categories. 

\noindent \textbf{COCO Ignore} We study the effectiveness COCO Ignore. 
As shown in Table \ref{tab:ablation}, COCO Ignore can significantly improve the $\mathrm{AP_{r}}$ by 4.7\% (10.1\% to 14.8\%), and lead an 0.7\% overall AP improvement. 

\noindent \textbf{COCO Pre-training} We demonstrate the effectiveness of COCO Pre-training in Table \ref{tab:ablation}. 
It brings consistent performance gain on three category groups.

\noindent \textbf{Resampling} We also implemented the class-aware data resampling method which is proposed in \cite{gupta2019lvisarvix}. 
We find that our equalization loss is compatible with resampling, EQL can further increase AP by 2\% with resampling (from 21.3\% to 23.3\%). 
Combining these two methods can improve the overall AP by 3.2\%.

\section{LVIS Challenge 2019}

We add several enhancements on our model for the challenge, results are shown in Table \ref{tab:enhancement}. 
With those enhancements, we achieve 36.4 and 28.9 Mask AP on \texttt{val} and \texttt{test} set respectively which is demonstrated in Table \ref{tab:testset}.

\noindent \textbf{Challenge Baseline} We replace ResNet50 with ResNeXt-101-64x-4d \cite{xie2017aggregated} and use synchronized batch normalization \cite{peng2018megdet} in backbone and heads.
Also deformable convolution is adopted \cite{dai2017deformable} in stage3, stage4 and stage5 of the model.
We use multi-scale training and images are resized to range from 400 to 1400, and the longer edge is set to 1400.
All these enhancements achieve an AP of 30.1\%, shown in Table~\ref{tab:enhancement}.

\noindent \textbf{Multi-scale Testing} We apply multi-scale testing on both bounding box and segmentation results. 
The testing scales are set to 600, 800, 1000, 1200, 1400.

\noindent \textbf{Expert Model} We train two large models on COCO and OpenImage respectively and do testing on LVIS test set. 
Shared categories detection results are used for ensemble. 
Expert models of COCO and OpenImage improve the AP by 0.2 and 0.2 respectively.

\noindent \textbf{Re-scoring Ensemble} Due to the class imbalance problem, the detection scores for rare and common categories are much lower than that of frequent categories. We observed that ensemble degrades recall of rare and common categories dramatically because of their low scores compared to frequent ones. To solve this, the predictions are sorted by their scores and category frequency jointly. For prediction $d_r$ of rare categories, and $d_f$ of frequent categories, $d_r$ is prior to $d_f$  if  $score_r + \alpha > score_f$, so do the common vs the frequent. In our experiments, we set $\alpha$ to 0.1 for the rare and 0.05 for the common, respectively.

\begin{table}
    \centering
    \small
    \begin{tabular}{l|c c c c}
         Model &$\mathrm{AP}$ & $\mathrm{AP_{r}}$ & $\mathrm{AP_{c}}$ & $\mathrm{AP_{f}}$  \\
        \hline 
        Challenge Baseline & 30.1 & 19.3 & 31.8 & 32.3 \\
        \hline
        +SE154 \cite{hu2018squeeze} & 30.8 & 19.7 & 32.2 & 33.4 \\
        +OpenImage Data & 31.4 & 21.5 & 33.1 & 33.3  \\
        +Multi-scale box testing & 32.3 & 20.5 & 34.7 & 34.2 \\
        +RS Ensemble + Expert Model & 35.1 & 24.8 & 37.5 & 36.3 \\
        +Multi-scale mask testing & \textbf{36.4} & \textbf{25.5}& \textbf{38.6} & \textbf{38.1} \\
         \hline
    \end{tabular}
    \caption{Experiment results of different tricks. RS Ensemble stands for Rescoring Ensemble.}
    \label{tab:enhancement}
\end{table}

\begin{table}
    \centering
    \begin{tabular}{l|c|c c c c}
          & eval.set & $\mathrm{AP}$ & $\mathrm{AP_{r}}$ & $\mathrm{AP_{c}}$ & $\mathrm{AP_{f}}$  \\
        \hline 
        \cite{gupta2019lvisarvix} & \texttt{val} & 27.1 & 15.6 & 27.5 & 31.4 \\
        Ours  & \texttt{val} & \textbf{36.4} & \textbf{25.5}& \textbf{38.6} & \textbf{38.1} \\
        \hline
        \cite{gupta2019lvisarvix} & \texttt{test} & 20.5 & 9.8 & 21.1 & 30.0 \\
        Ours & \texttt{test} & \textbf{28.9} & \textbf{17.7} & \textbf{30.8} & \textbf{36.7} \\
         \hline
    \end{tabular}
    \caption{Final results on val and test.}
    \label{tab:testset}
\end{table}

{\small \bibliographystyle{ieee_fullname} \bibliography{egbib}}

\end{document}